\documentclass[letterpaper, 10 pt, conference]{ieeeconf}  
\IEEEoverridecommandlockouts
\usepackage{helvet}         		% selects Helvetica as sans-serif font
\usepackage{type1cm}        		% activate if the above 3 fonts are not available on your system
\usepackage{graphics} 		% for pdf, bitmapped graphics files
\usepackage{graphicx}        % standard LaTeX graphics tool when including figure files
\usepackage{epsfig} 			% for postscript graphics files
\usepackage{multicol}        		% used for the two-column index
\usepackage[bottom]{footmisc}	% places footnotes at page bottom

\usepackage{amssymb}
\usepackage[ruled]{algorithm2e}
\usepackage{subfigure}
\usepackage{newtxtext}       % 
\usepackage{newtxmath}       % selects Times Roman as basic font
\usepackage{caption}
\usepackage{gensymb}
\usepackage{booktabs}
\usepackage[table,xcdraw]{xcolor}
\usepackage{epstopdf}
\usepackage{url}
\usepackage[font={small}]{caption}
\usepackage{commath}

\DeclareMathAlphabet{\mathcal}{OMS}{cmmi}{m}{n}

\DeclareFontFamily{U}{mathc}{}
\DeclareFontShape{U}{mathc}{m}{it}%
{<->s*[1.03] mathc10}{}
\DeclareMathAlphabet{\mathcal}{U}{mathc}{m}{it}

\newcommand{\fbf}{\mathbf{f}}

\newcommand{\lbf}{\mathbf{l}}

\newcommand{\sbf}{\mathbf{s}}

\newcommand{\xbf}{\mathbf{x}}

\newcommand{\Ebf}{\mathbf{E}}

\newcommand{\Bcal}{\mathcal{B}}

\newcommand{\Dcal}{\mathcal{D}}

\newcommand{\Fcal}{\mathcal{F}}
\newcommand{\Gcal}{\mathcal{G}}

\newcommand{\Lcal}{\mathcal{L}}

\newcommand{\Rcal}{\mathcal{R}}

\newcommand{\Ucal}{\mathcal{U}}

\newcommand{\Xcal}{\mathcal{X}}

\newcommand{\Fsf}{\mathsf{F}}
\newcommand{\Gsf}{\mathsf{G}}

\title{\LARGE \bf
Zero-Shot Reinforcement Learning on Graphs for Autonomous Exploration Under Uncertainty %Active SLAM Using A Single Training Environment
}

\author{Fanfei Chen, Paul Szenher, Yewei Huang, Jinkun Wang, Tixiao Shan, Shi Bai and Brendan Englot 
\thanks{F. Chen, P. Szenher, Y. Huang, J. Wang and B. Englot are with the Department of Mechanical Engineering, Stevens Institute of Technology, USA, 
        {\tt\small \{fchen7, pszenher, yhuang85, jwang92, benglot\}@stevens.edu}}

\thanks{T. Shan is with the Computer Science \& Artificial Intelligence Laboratory, Massachusetts Institute of Technology, USA, 
        {\tt\small shant@mit.edu}}

\thanks{S. Bai is with Wing, Alphabet Inc.,
        {\tt\small baishi@wing.com}}
}

\begin{document}

\maketitle
\thispagestyle{empty}
\pagestyle{empty}

%%%%%%%%%%%%%%%%%%%%%%%%%%%%%%%%%%%%%%%%%%%%%%%%%%%%%%%%%%%%%%%%%%%%%%%%%%%%%%%%%%%%%%%%%
%%%%%%%%%%%%%%%%%%%%%%%%%%%%%%%%%%%%%%%%%%%%%%%%%%%%%%%%%%%%%%%%%%%%%%%%%%%%%%%%%%%%%%%%%
\begin{abstract}

This paper studies the problem of autonomous exploration under localization uncertainty for a  mobile robot with 3D range sensing.
We present a %learning-based active SLAM 
framework for self-learning a high-performance exploration policy in a single simulation environment, and transferring it to other environments, which may be physical or virtual. Recent work in transfer learning achieves encouraging performance by domain adaptation and domain randomization to expose an agent to scenarios that fill the inherent gaps in sim2sim and sim2real approaches. However, it is inefficient to train an agent in environments with randomized conditions to learn the important features of its current state. An agent can use domain knowledge provided by human experts to learn efficiently. We propose a novel approach that uses graph neural networks in conjunction with deep reinforcement learning, enabling decision-making over graphs containing relevant exploration information provided by human experts to predict a robot’s optimal sensing action in belief space. The policy, which is trained only in a single simulation environment, offers a real-time, scalable, and transferable decision-making strategy, resulting in zero-shot transfer to other simulation environments and even real-world environments.

\end{abstract}
\label{sec:1}
%%%%%%%%%%%%%%%%%%%%%%%%%%%%%%%%%%%%%%%%%%%%%%%%%%%%%%%%%%%%%%%%%%%%%%%%%%%%%%%%%%%%%%%%%
%%%%%%%%%%%%%%%%%%%%%%%%%%%%%%%%%%%%%%%%%%%%%%%%%%%%%%%%%%%%%%%%%%%%%%%%%%%%%%%%%%%%%%%%%
\section{Introduction}
In this paper we consider the autonomous exploration of an unknown environment by a range-sensing mobile robot reliant upon simultaneous localization and mapping (SLAM).
%This a challenging variant of active simultaneous localization and mapping (SLAM).
%Active Simultaneous localization and mapping (SLAM) of a priori unknown environments is a challenging problem.
Many recent solutions to this variant of active SLAM adopt a utility function that manages the trade-off between exploration and place-revisiting, and have high computational complexity due to the need for forward-simulation of SLAM along candidate paths. Although learning-based exploration algorithms can estimate an optimal action with greatly reduced %nearly constant 
computation time, these algorithms can require extensive training across many representative environments, and are vulnerable to poor generalizability when transferring a learned policy to a new %unknown 
environment. Our recent prior work \cite{Chen2020} proposed a generalized graph representation, which we termed the \textit{exploration graph}, for learning-based exploration under uncertainty, compatible with 2D landmark-based SLAM, deep reinforcement learning (DRL) \cite{Mnih2015} and graph neural networks (GNNs). However, many complex real-world environments cannot be captured using landmarks, and this approach has limited applicability in such environments.
%it is difficult to define the landmarks in the SLAM graph so it is challenging to implement an exploration graph in a real-world environment.

\begin{figure}[t]
\centering
\includegraphics[width=0.99\columnwidth]{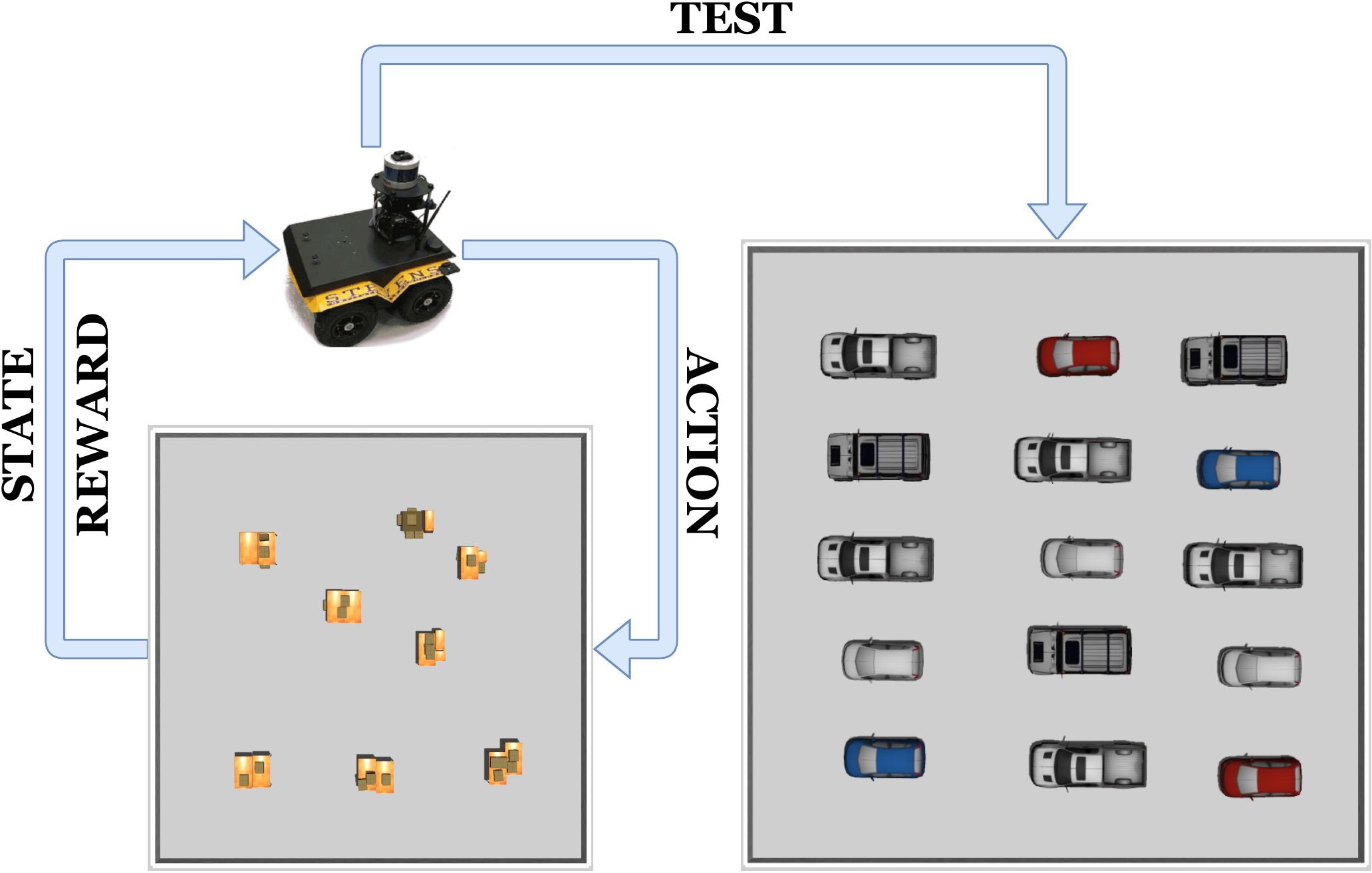}
\caption{\textbf{An illustration of our framework.} The robot is trained in a single representative simulation environment by DRL with GNNs. %The robot has nine fixed different initial locations in the training environment. 
Then the GNN-based policy guides the robot to perform its exploration task in a testing environment of a different size, containing new objects, which may be real or virtual. %of different shapes and sizes.
}
\label{framkework}
\vspace{-5mm}
\end{figure}

We propose a new structure for the exploration graph in this paper, intended to support pose SLAM with dense 3D observations; the graph
incorporates only poses and frontiers as nodes, without landmarks, to provide more generalizability. With the feature vector for the GNN provided by human experts, the agent has access to parameters directly relevant to the active SLAM task. Therefore the learning agent in our framework, despite the complexity of its dense 3D observations, only needs to use a single training environment to learn, from graphs, how to manage the trade-off between exploration and place revisiting. Fig. \ref{framkework} summarizes our approach. %The robot is trained in an office-like Gazebo environment, initialized from different locations. % with nine different robot initial locations. 
After training in a virtual environment, the robot will use its learned policy without fine-tuning to explore new unknown (real and virtual) environments containing obstacles of different sizes, shapes, and spatial arrangements.

In this paper, we present a learning-based active SLAM framework with DRL on an optimized exploration graph which only needs a single virtual training environment to perform zero-shot RL transfer to other real and virtual environments\footnote{Video attachment: \url{https://youtu.be/62phOSf2HEg}}. We demonstrate that the GNN-based RL policy can be trained efficiently and transferred to new environments without parameter-tuning.

%%%%%%%%%%%%%%%%%%%%%%%%%%%%%%%%%%%%%%%%%%%%%
\subsection{Related Work}
Information-theoretic exploration methods focus on efficiently reducing the entropy in a mobile robot's evolving occupancy map.
Specifically, \cite{Julian2014} repeatedly selects the robot sensing action that maximizes the mutual information (MI) between the robot's range beams and the cells of its occupancy map. The Cauchy-Schwarz quadratic mutual information (CSQMI) has also been adopted, to reduce the computation time during sensing action selection \cite{Charrow2015CSQMI}. \cite{Jadidi2018} proposed to travel to the most informative map frontier predicted using Gaussian process (GP) occupancy maps.

Occupancy map entropy and a robot's localization uncertainty are both considered in active SLAM exploration algorithms. The entropy of the robot trajectory and its map are utilized in \cite{Valencia2012} to reduce both quantities. In \cite{Stachniss2005}, robot uncertainty during exploration dictates the particle weights applied in a particle filtering framework. Wang et al. \cite{Wang2017}, \cite{Wang2019} proposed using \textit{virtual landmarks} within an Expectation-Maximization (EM) inspired exploration algorithm to manage both robot uncertainty and its influence on occupancy map accuracy, using a novel utility function that explores while driving down the total uncertainty of a \textit{virtual map}.

However, the high computational cost of the above state-of-the-art approaches limits their %use in real-time applications. 
scalability.
Alternatively, learning-based exploration methods can provide scalable, real-time decision making and near-optimal exploration policies. Without considering uncertainty, \cite{Bai2015}, \cite{Bai2016} adopted GP modeling of MI and a subsequent Bayesian optimization active sampling approach to reduce the computational complexity of information-theoretic exploration. Furthermore, a supervised deep learning method \cite{Bai2017} was proposed to predict the most informative sensing action given a local submap as input. DRL has also been used to select near-optimal, entropy-reducing sensing actions using 2D \cite{Niroui2019} and 3D \cite{Chen2019} occupancy maps as input data. However, such learning-based methods need to capture a large variety of parameters to be applied successfully in complex environments. The training processes are intensive, and the learned policy is often challenging to transfer to a new environment. 

Graphs can provide a generalized topological representation of a robot and its environment, and learning from graphs is an emerging research area. Combining graphs with neural network models \cite{Scarselli2009} provides exceptional performance in many research fields. Graph Nets \cite{Battaglia2018} are adopted to solve control problems by formulating a graph that represents the state of a dynamical system \cite{Sanchez2018}. A robot navigation problem is solved in \cite{KChen2019} with localization graphs and camera images. Combining graphs with DRL, Wang et al. \cite{Wang2018} showed policies learned with GNNs are capable of solving control problems. To address the transfer of policies learned for mobile robot exploration under localization uncertainty, our prior work proposed adopting an \textit{exploration graph} as a generalized representation of a robot's state \cite{Chen2020}, \cite{Chen2019ISRR}. 

Transfer learning offers a mechanism for training robots safely in simulation environments, avoiding the expense of training a complex system completely in real-world environments. Bruce et al. \cite{Bruce2017} proposed an off-line interactive replay for real-world environment DRL training. In \cite{Mordatch2015}, domain randomization is adopted to fill the gap between simulated and real-world environments. Genc et al. \cite{Genc2019} proposed zero-shot reinforcement learning using an attention mechanism to extract important features from camera images.

In this work we revisit the \textit{exploration graph} as an appealing generalized representation of a robot's state, trajectory and environment, which, when combined with relevant features selected by human experts, allows learned exploration policies to be successfully transferred to new environments. %Our proposed approach, detailed in the sections to follow, offers efficient training and a robust transferable learned policy. 
The new measures described in the sections below specifically address how our DRL GNN framework can be applied, for the first time, to the exploration of environments populated with complex obstacles under dense 3D sensor observations. Successful training now relies on high-fidelity simulation, and accordingly, our approach offers a highly efficient training process to meet these new requirements. 

%%%%%%%%%%%%%%%%%%%%%%%%%%%%%%%%%%%%%%%%%%%%%
\subsection{Paper Organization}
We first introduce a SLAM framework compatible with the proposed exploration graph in Section \ref{sec:Formulation}. We then explain the framework for learning a robot exploration policy in Section \ref{sec:Algorithms}. In Section \ref{sec:experiments}, we present the testing results of our robot exploration experiments, with conclusions provided in Section \ref{sec:conclusions}.

\begin{figure*}[ht]
\centering
\subfigure[]{\includegraphics[width=0.47\textwidth]{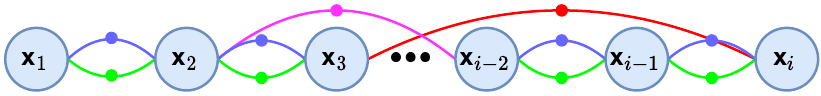}\label{eg1}}
\subfigure[]{\includegraphics[width=0.47\textwidth]{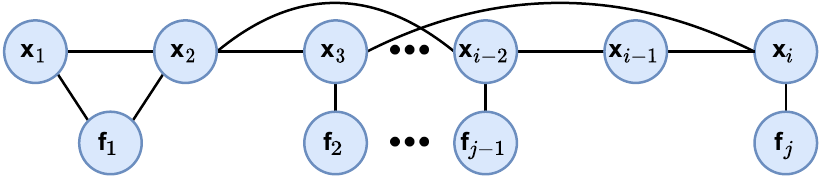}\label{eg2}}
\vspace{-1mm}
\caption{\textbf{An illustration of the SLAM factor graph and the exploration graph.} Left: The SLAM factor graph contains four different types of constraints: \textcolor{blue}{blue factors} are provided by odometry measurements; \textcolor{green}{green factors} are obtained from sequential scan matching of two consecutive poses; \textcolor{red}{red factors} represent the loop closures provided by point cloud segment matching; \textcolor{magenta}{magenta factors} are loop closures generated by pose matching.
Right: The corresponding exploration graph is shown, containing all poses from the SLAM factor graph, and waypoints representing map frontiers. If two poses have a constraint joining them in the SLAM factor graph, an edge will be assigned to connect these two poses. The current pose $x_i$ is connected to the nearest frontier, and any frontiers whose paths achieve place revisiting are connected to the prior poses they revisit. All the edges in the exploration graph are weighted with Euclidean distances.}
\label{graphs}
\vspace{-5mm}
\end{figure*}

%%%%%%%%%%%%%%%%%%%%%%%%%%%%%%%%%%%%%%%%%%%%%%%%%%%%%%%%%%%%%%%%%%%%%%%%%%%%%%%%%%%%%%%%%
%%%%%%%%%%%%%%%%%%%%%%%%%%%%%%%%%%%%%%%%%%%%%%%%%%%%%%%%%%%%%%%%%%%%%%%%%%%%%%%%%%%%%%%%%
\section{Problem Formulation and Approach}
\label{sec:Formulation}
%%%%%%%%%%%%%%%%%%%%%%%%%%%%%%%%%%%%%%%%%%%%%
\subsection{Simultaneous Localization and Mapping Framework}
We use the same SLAM framework as in \cite{Wang2019} to support exploration. An illustration of the SLAM factor graph is shown in Fig. \ref{eg1}. There are two types of sequential factors; blue factors indicate the \textit{odometry} measurements ($\mathbf \phi^O$) between two consecutive poses, and green factors represent \textit{sequential scan matching} constraints ($\mathbf \phi^{\text{SSM}}$). These are provided by the iterative closest point (ICP) algorithm, using 3D LiDAR point clouds to estimate the relative transformation between consecutive poses. There are also two types of loop closure constraints. \textit{Pose matching} ($\mathbf \phi^{\text{PM}}$), indicated by magenta in Fig. \ref{eg1}, provides a loop closure constraint when the point cloud from the current pose has been successfully matched with the point cloud from a previous pose. Finally, \textit{segment matching} ($\mathbf \phi^{\text{SM}}$), colored by red in Fig. \ref{eg1}, achieves a loop closure when two poses observe the same segmented object in their respective point clouds (matched using descriptors, according to the approach of \cite{segmatch}).

The motion model and the measurement model of our SLAM framework are defined as:
\begin{align}
& \mathbf x_i = h_i(\mathbf x_{i-1}, \mathbf u_i) + \mathbf w_i,\quad \mathbf w_i \sim \mathcal N(\mathbf 0, Q_i),\\
& \mathbf z_{k} = g_{k}(\mathbf x_{i_k}) + \mathbf v_{k},\quad \mathbf v_{k} \sim \mathcal N(\mathbf 0, R_{k}),
\end{align}
where $\mathcal X = \{\mathbf x_i\}_{i=1}^t$ are 6-DOF robot poses and $\mathcal{U}=\{ \mathbf{u_i} \}_{i=1}^t$ is a given motion input. Then we define the factor graph:
\begin{flalign*}
\mathbf \phi(\mathbf x) = \mathbf  \phi^{\text{0}}(\mathbf x_0) & \prod_i \mathbf \phi^{\text{O}}_{i}(\mathbf x_i) \prod_j \mathbf \phi^{\text{SSM}}_j(\mathbf x_j) & \text{(sequential)}\\
&\prod_p \mathbf \phi^{\text{PM}}_p(\mathbf x_p)\prod_q \mathbf \phi^{\text{SM}}_q(\mathbf x_q). &  \text{(loop closures)}
\end{flalign*}
The SLAM problem then becomes a nonlinear least-squares optimization problem on a factor graph. We use iSAM2 \cite{Kaess2012} to solve this problem. 

We adopt the \textit{virtual map} framework from the EM exploration algorithm of \cite{Wang2017}. A virtual map is uniformly discretized at the same or lower resolution than the robot's occupancy map, and contains \textit{virtual landmarks}, $\tilde{\lbf}_k \in \tilde{\Lcal}$, populating the map's cells. Each virtual landmark has a large initial covariance, which will be driven down by the robot as it observes the contents of the map cell. In this setting, the goal of exploration is to minimize the covariance of all \textit{virtual landmarks}, which leads a robot to both explore efficiently and to produce an accurate map. The utility function of the current state is defined as follows:
\begin{equation}
\label{utility_function}
U(\tilde{\Lcal}) = \sum_{\tilde{\lbf}_k \in \tilde{\Lcal}} \log \text{det} (\Sigma_{\tilde{\lbf}_k}),
\end{equation}
where $\Sigma_{\tilde{\lbf}_k}$ is the covariance matrix for virtual landmark $\tilde{\lbf}_k$.

\vspace{-2mm}
%%%%%%%%%%%%%%%%%%%%%%%%%%%%%%%%%%%%%%%%%%%%%
\subsection{Exploration Graph}
\vspace{-1mm}
We define the exploration graph $\mathcal{G}=\mathcal{(V, E)}$ as shown in Fig. \ref{eg2}. There are two types of vertices in $\mathcal{V}$. $\Xcal \subset \mathcal{V}$ contains the robot pose history. Poses connected by constraints in the SLAM factor graph in Fig. \ref{eg1} are also connected with edges in the exploration graph. Furthermore, the exploration graph includes exploration waypoints derived from map frontiers, $\Fcal \subset \mathcal{V}$. These frontier nodes are extracted from the map's boundaries between free and unknown areas. %In our exploration framework, we only consider frontiers which either provide place-revisiting to reduce the uncertainty of the current state, or allow for exploration of unknown space in the current occupancy map. Hence,
The current pose $x_t$ is connected with the \textit{nearest} frontier to its location. If a frontier can provide place-revisiting through either pose matching or segment matching, we connect the previous poses associated with that loop closure to this frontier. All other frontiers, which are neither the nearest frontier to the current pose, nor achieve place-revisiting, are excluded from the exploration graph. All edges in the graph are weighted by their Euclidean distances.

Each vertex $\mathbf{n_i} \in \mathcal V$ has a feature vector
\begin{align}
& \sbf_i=[s_{i_1}, s_{i_2}, s_{i_3}, s_{i_4}],\nonumber\\
& s_{i_1}= \phi_\text{A}(\Sigma_i), \label{s1} \\
& s_{i_2}= \sqrt{{(x_i-x_t)}^2 + {(y_i-y_t)}^2}, \label{s2} \\
& s_{i_3}= \text{arctan2}(y_i-y_t,x_i-x_t), \label{s3} \\
& s_{i_4}= 
        \begin{cases}
            0 & \mathbf{n_{i}} = \mathbf{x_t} \\
            1 & \mathbf{n_{i}} \in \{ \mathbf{f_n} \} \\
            -1 & \text{otherwise}
        \end{cases}. \label{s4}
\end{align}
In Eq. (\ref{s1}), we adopt the A-Optimality \cite{Kaess2009} criterion as a metric to evaluate the uncertainty level of each node. The covariances of frontier vertexes $\Fcal$ are extracted from the virtual map. We capture the geometric information of the current robot state using Eqs. (\ref{s2}) and (\ref{s3}), which contain the relative distance and orientation information between current pose $x_t$ and the node $n_i$. The last feature $s_{i_4}$ is used to indicate the identity of each node. The current pose is labeled as 0, all previous poses are -1, and frontiers are 1.

\vspace{-1.5mm}

\section{Algorithms and System Architecture}
\label{sec:Algorithms}
%%%%%%%%%%%%%%%%%%%%%%%%%%%%%%%%%%%%%%%%%%%%%
\subsection{Graph Neural Networks}

In this paper, we use two Graph U-Nets (g-U-Nets) \cite{Gao2019} to serve as the policy network and the value network, respectively. Similar to U-Net \cite{Ronneberger2015}, g-U-Nets have graph pooling layers to encode the input feature vectors of nodes in the input graph. Additionally, the graph unpooling layers are used to decode the graphs in the hidden layers to provide the output graphs. Besides the pooling and unpooling layers, each encoder and decoder has a Graph Convolutional Network (GCN) \cite{Kipf2017} layer to update the graph features. The depth of our g-U-Nets is 3 and the number of features for each hidden layer is 1000. A multilayer perceptron (MLP) output layer is adopted to provide the final output prediction. A dropout layer with a 0.5 dropout rate is placed between the output of our g-U-Nets and the MLP output layer.

\vspace{-1.5mm}

%%%%%%%%%%%%%%%%%%%%%%%%%%%%%%%%%%%%%%%%%%%%%
\subsection{Deep Reinforcement Learning}

In our framework, the robot is solving a decision-making problem over exploration graphs using a learned policy from DRL. The candidate actions are paths from the current pose to the frontier nodes in the exploration graph. The exploration graph contains information about all past poses, their respective uncertainty, and how they relate to map frontiers. At each decision-making instant $k$, the state of our system is represented by an exploration graph $\Gcal_k\in\Gsf$. A reward $R_k\in\mathbb{R}$ is assigned to the selected action $\fbf_k\in\Fsf_{\Gcal_k}$. The overall decision-making process can be modeled as a Markov Decision Process $\left<\Gsf, \Fsf, \textnormal{Pr}, \gamma \right>$ \cite{Puterman1994}. 

\begin{algorithm}[t]
\caption{Reward Function}
\label{alg:reward}
\textbf{input:} Exploration graph $\Gcal$, Frontier node $\fbf$\\
{\color{gray} \# Calculate the raw reward (Eq. \ref{raw_reward_function})}\\
$\Rcal^0_{\Gcal}=U^0(\tilde{\Lcal}) - U_\Ucal'(\tilde{\Lcal}')- \alpha C(\Ucal)$ \\
{\color{gray} \# Normalize the raw reward}\\
$\Rcal_{\Gcal} = \{ r_{\fbf'} = \Rcal^0_{\Gcal}\}$\\
$l = \min\Rcal_{\Gcal}$, $u = \max\Rcal_{\Gcal}$\\
$r_{\fbf} \gets (r_{\fbf} - l)/(u-l)$\\
{\color{gray} \# Compute projection based on nearest frontier}\\
$\fbf_t=\texttt{nearest\_frontier}(\xbf_t)$\\
\If{$u$ = \texttt{raw\_reward}$(\fbf_t)$}{
    \textbf{return} $r_{\fbf}-1$ {\color{gray} \# $r(\Gcal,\fbf)\in[-1,0]$}
}
\textbf{return} $2r_{\fbf}-1$ {\color{gray} \# $r(\Gcal,\fbf)\in[-1,1]$}
\end{algorithm}

\vspace{-0mm}

We consider the A2C \cite{Vinyals2017} policy-based DRL algorithm in this paper, for which two separate g-U-Nets serve as the policy network and the value network. The loss function is defined as follows:
\begin{align}
\label{a2c_loss}
L_{\textnormal{A2C}}(\Dcal)&=\Ebf_{\Bcal\sim\Dcal}[L_{\textnormal{A2C}}^{(1)}+\eta L_{\textnormal{A2C}}^{(2)}],\\
L_{\textnormal{A2C}}^{(1)} &= [A(\Gcal,\fbf)\log\pi(\fbf|\Gcal)+\beta(A(\Gcal,\fbf))]^2 \nonumber, \\ 
L_{\textnormal{A2C}}^{(2)} &= \sum_{\fbf \in \Fsf_{\Gcal}}\pi(\fbf|\Gcal)\log\pi(\fbf|\Gcal)\nonumber,
\end{align}
where the \textit{advantage} function is defined as $A(\Gcal,\fbf) = Q(\Gcal,\fbf)-V(\Gcal)$ to evaluate the difference between the state value and the state-action value. $\beta\in\mathbb{R}$ is a coefficient for the loss of the value function. The entropy coefficient $\eta\in\mathbb{R}^+$ is used to weigh output entropy for encouraging exploration during training.

%%%%%%%%%%%%%%%%%%%%%%%%%%%%%%%%%%%%%%%%%%%%%
\subsection{Reward Function}
We use the utility function given in Eq. (\ref{utility_function}) from the EM exploration algorithm \cite{Wang2017}, whose behavior we wish to emulate, to compute the raw reward for actions that travel to each candidate frontier. The raw reward is defined as follows:
\begin{equation}
\label{raw_reward_function}
\Rcal^0_{\Gcal} = U^0(\tilde{\Lcal}) - U_\Ucal'(\tilde{\Lcal}')- \alpha C(\Ucal),
\end{equation}
where $\Ucal$ contains sequential actions to the selected frontier position. The output of cost-to-go function $C(\Ucal)$ is the travel distance weighed by coefficient $\alpha$, expressing a preference for shorter paths. We then set a new range for these raw rewards by linear normalization. If the frontier selected by the EM algorithm is the \textit{nearest} frontier associated with the current pose, the range of the reward is $[-1,0]$. Otherwise, the range is $[-1,1]$. The reward function is described in Alg. \ref{alg:reward}.

\vspace{2mm}

%%%%%%%%%%%%%%%%%%%%%%%%%%%%%%%%%%%%%%%%%%%%%
\subsection{Computational Complexity}

For the EM algorithm, the computational complexity of the decision-making process is $\mathcal O(N_{\text{actions}}(C_1 + C_2))$, where %\cite{Wang2017}
$C_1=\mathcal O(n^3)$ bounds the complexity of the iSAM2 update ($n$ is the number of poses), %\cite{Kaess2012}
and $C_2=\mathcal O(m)$ bounds the covariance update for virtual landmarks ($m$ is the number of virtual landmarks) \cite{Wang2017}. The computation time increases significantly in the size of the state-action space, limiting the framework's applicability. On the other hand, as shown in \cite{Chen2020}, the computation time for decision-making with our fully trained DRL GNN framework is nearly constant, allowing real-time performance across a wide range of problems. 

%%%%%%%%%%%%%%%%%%%%%%%%%%%%%%%%%%%%%%%%%%%%%%%%%%%%%%%%%%%%%%%%%%%%%%%%%%%%%%%%%%%%%%%%%
%%%%%%%%%%%%%%%%%%%%%%%%%%%%%%%%%%%%%%%%%%%%%%%%%%%%%%%%%%%%%%%%%%%%%%%%%%%%%%%%%%%%%%%%%
\section{Experiments and Results}
\label{sec:experiments} 

\begin{figure}[t]
\centering
\includegraphics[height=38mm]{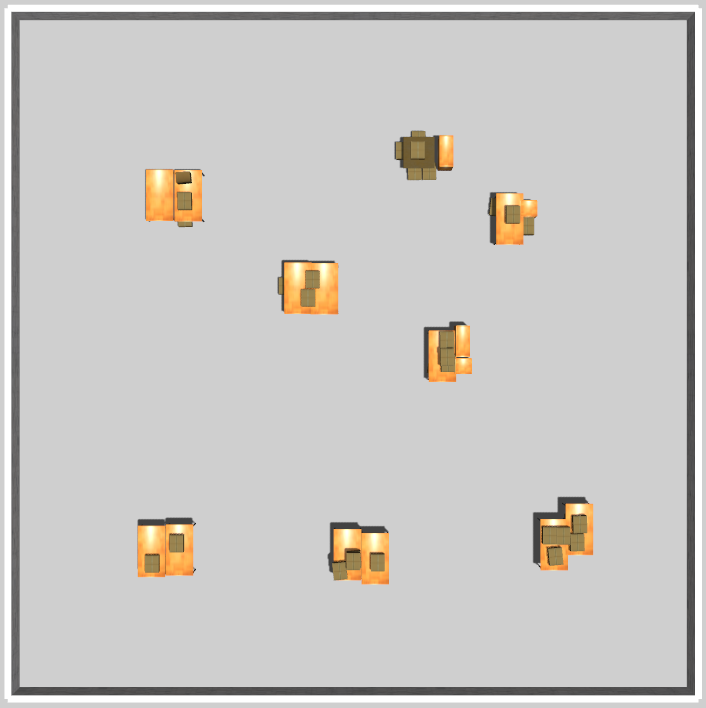}
\caption{The office-like Gazebo environment used for training.}
\label{train_env}
\vspace{-0mm}
\end{figure}

\begin{figure}[t]
\centering
\includegraphics[height=38mm]{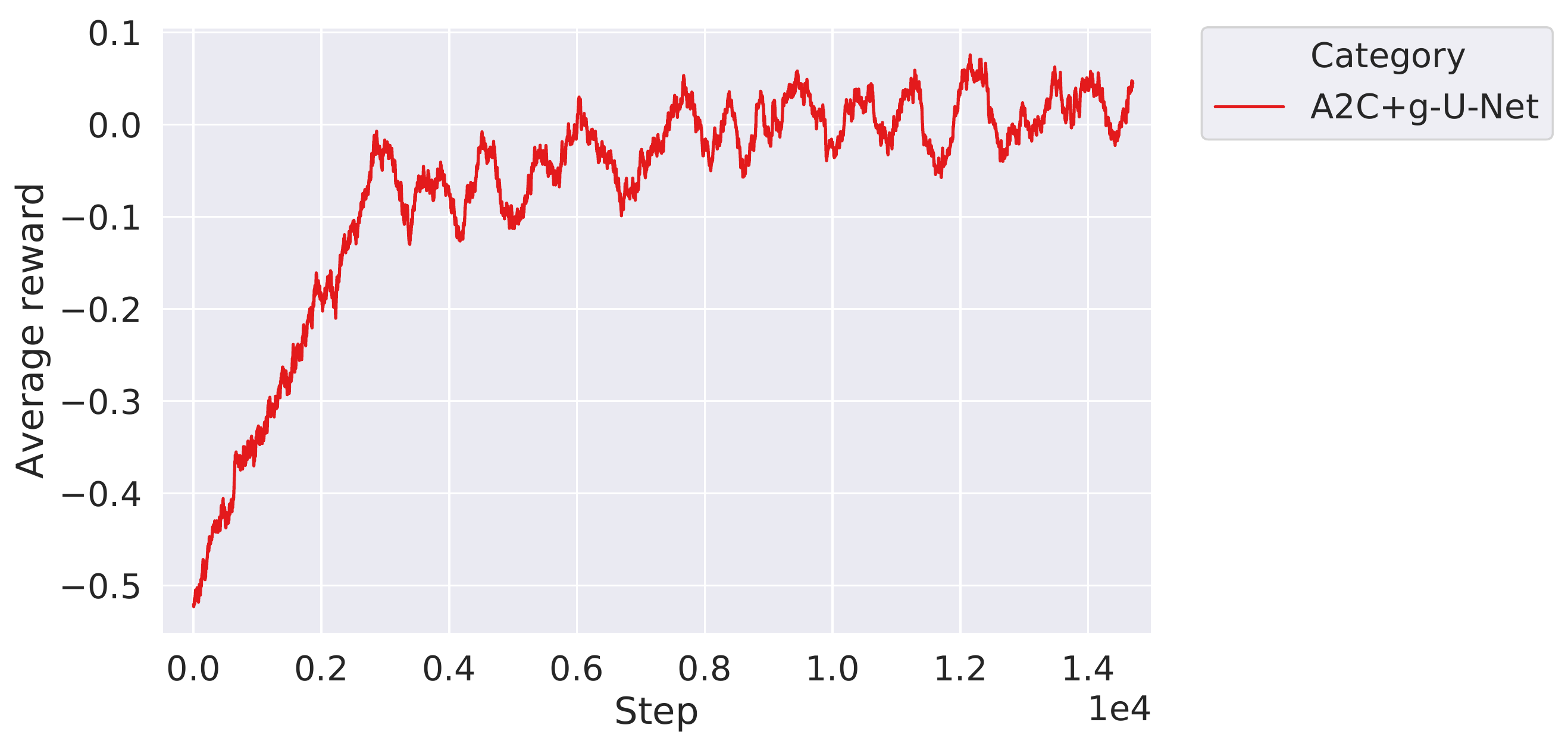}
\caption{The average reward during training.}
\label{ave_reward}
\vspace{-6mm}
\end{figure}

%%%%%%%%%%%%%%%%%%%%%%%%%%%%%%%%%%%%%%%%%%%%%
\subsection{Experimental Setup}

\begin{figure*}[t]
\raggedleft
\subfigure[]{\includegraphics[height=36mm]{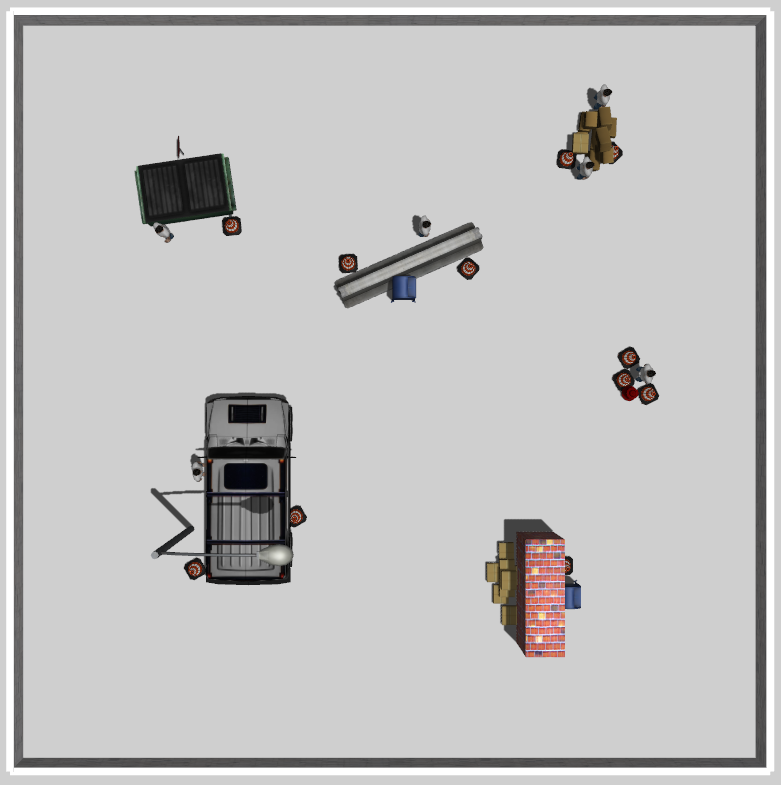}\label{test_left_env}}
\subfigure[]{\includegraphics[height=36mm]{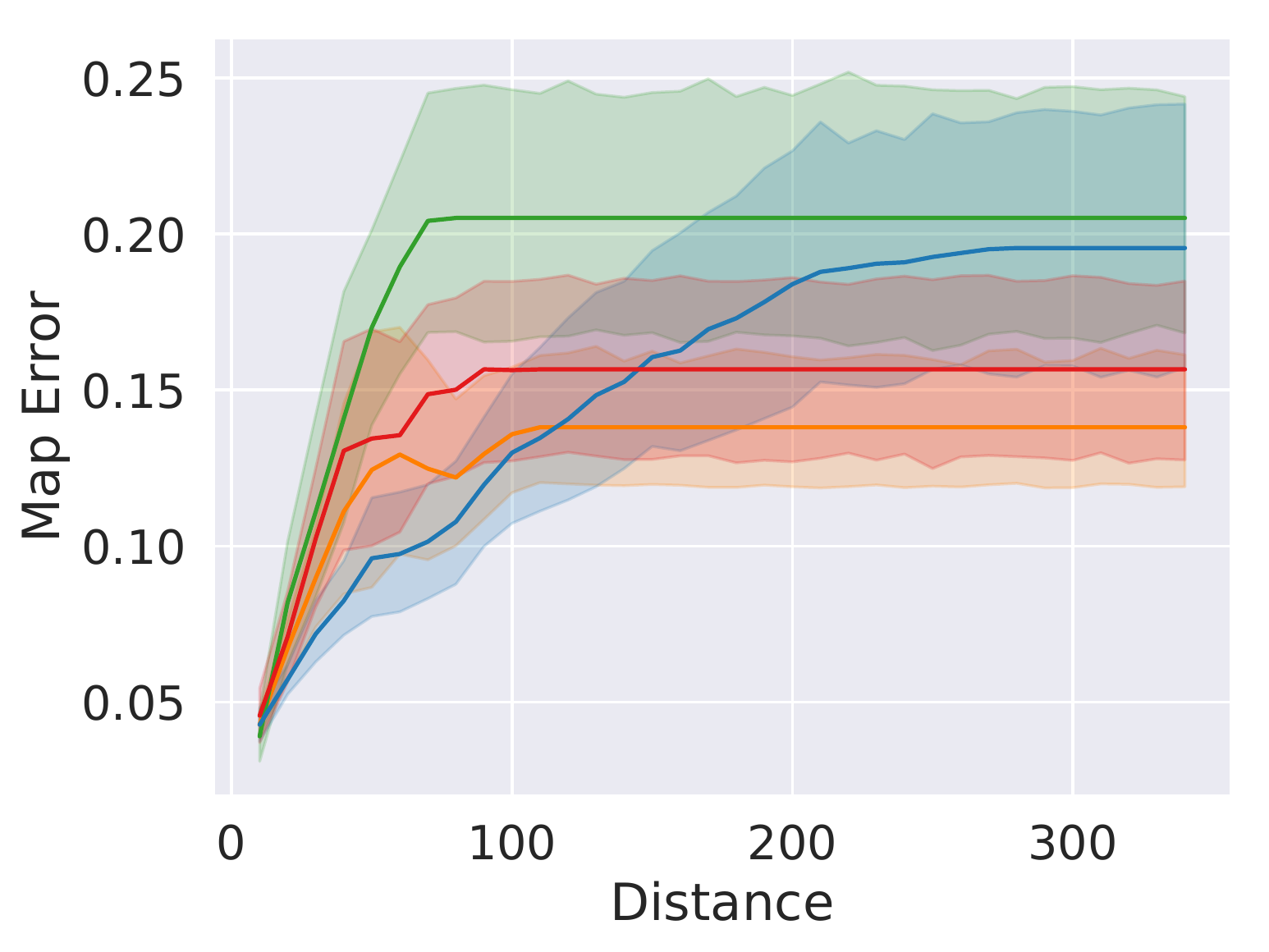}\label{test_left_error}}
\subfigure[]{\includegraphics[height=36mm]{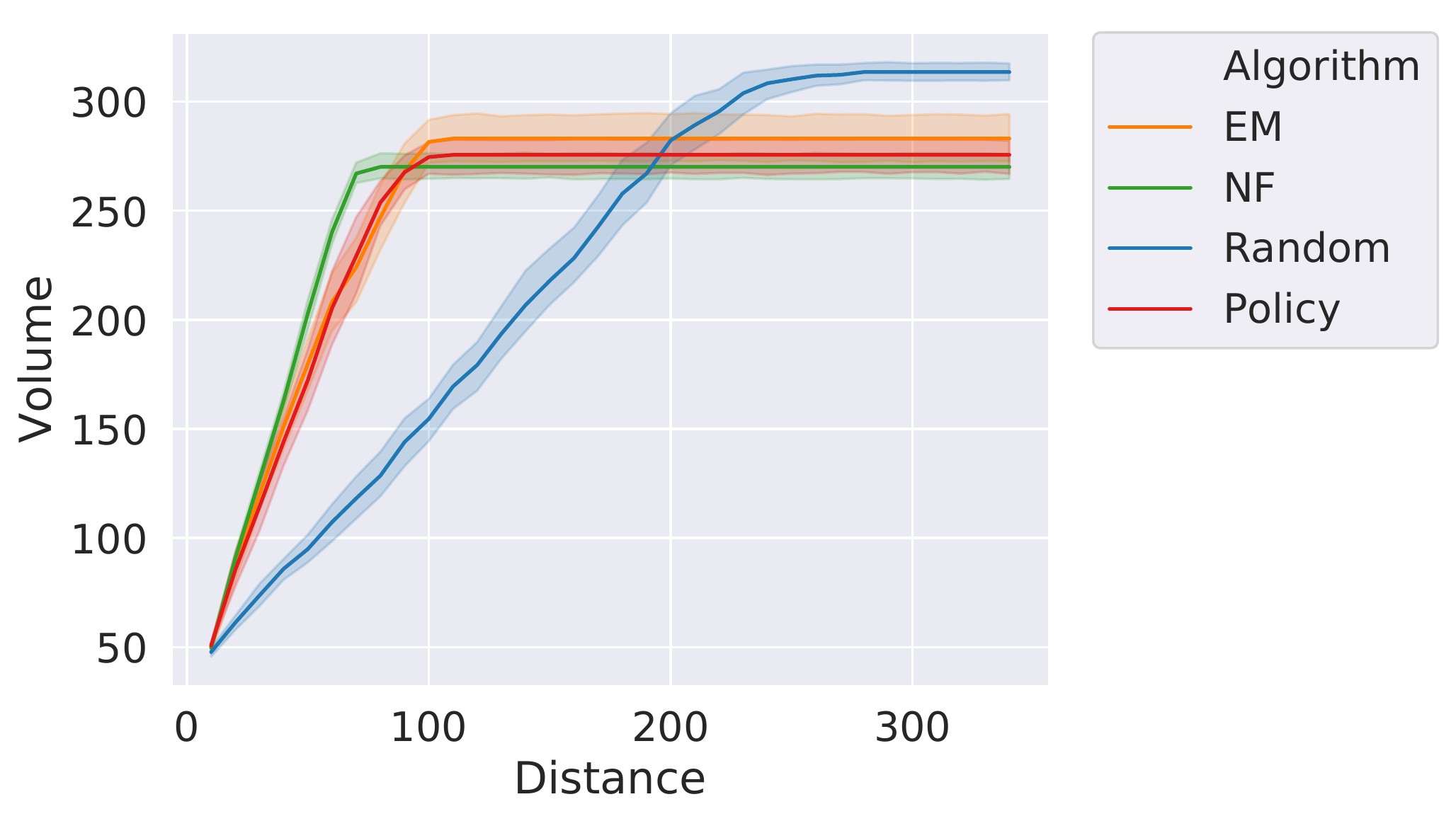}\label{test_left_volume}}
\subfigure[]{\includegraphics[height=55mm]{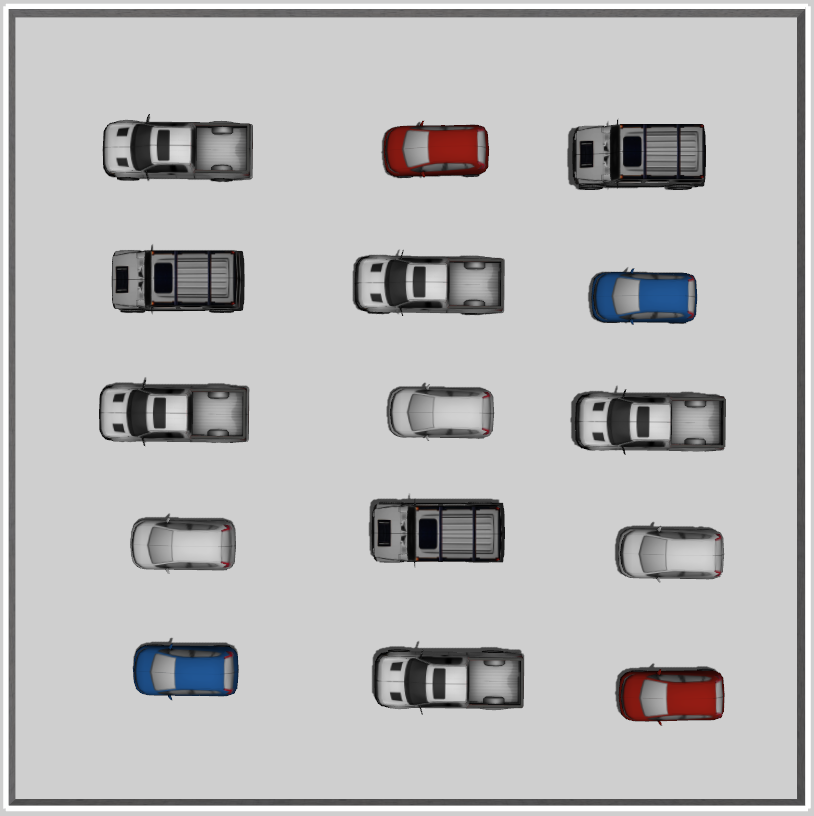}\label{test_right_env}}
\subfigure[]{\includegraphics[height=36mm]{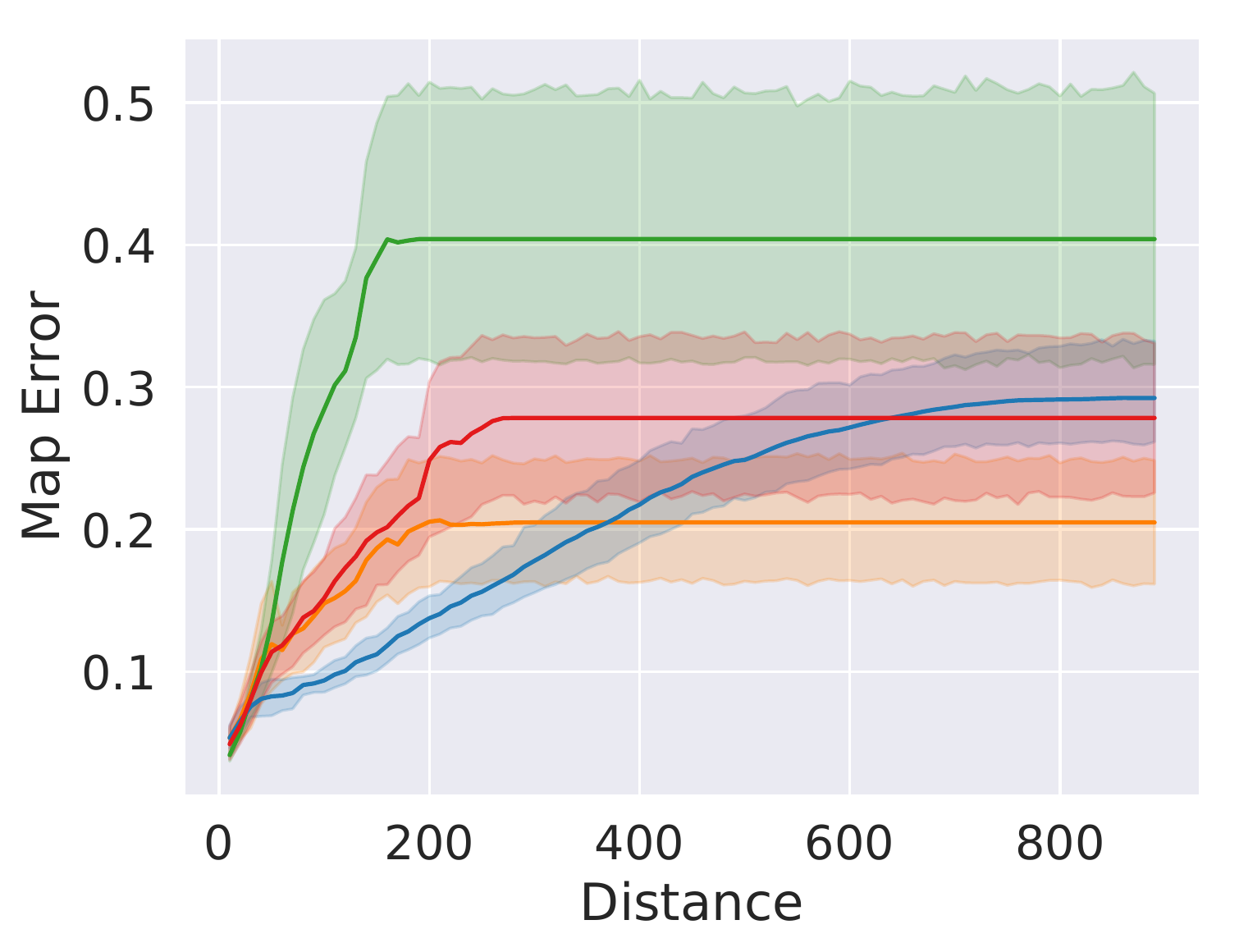}\label{test_right_error}}
\subfigure[]{\includegraphics[height=36mm]{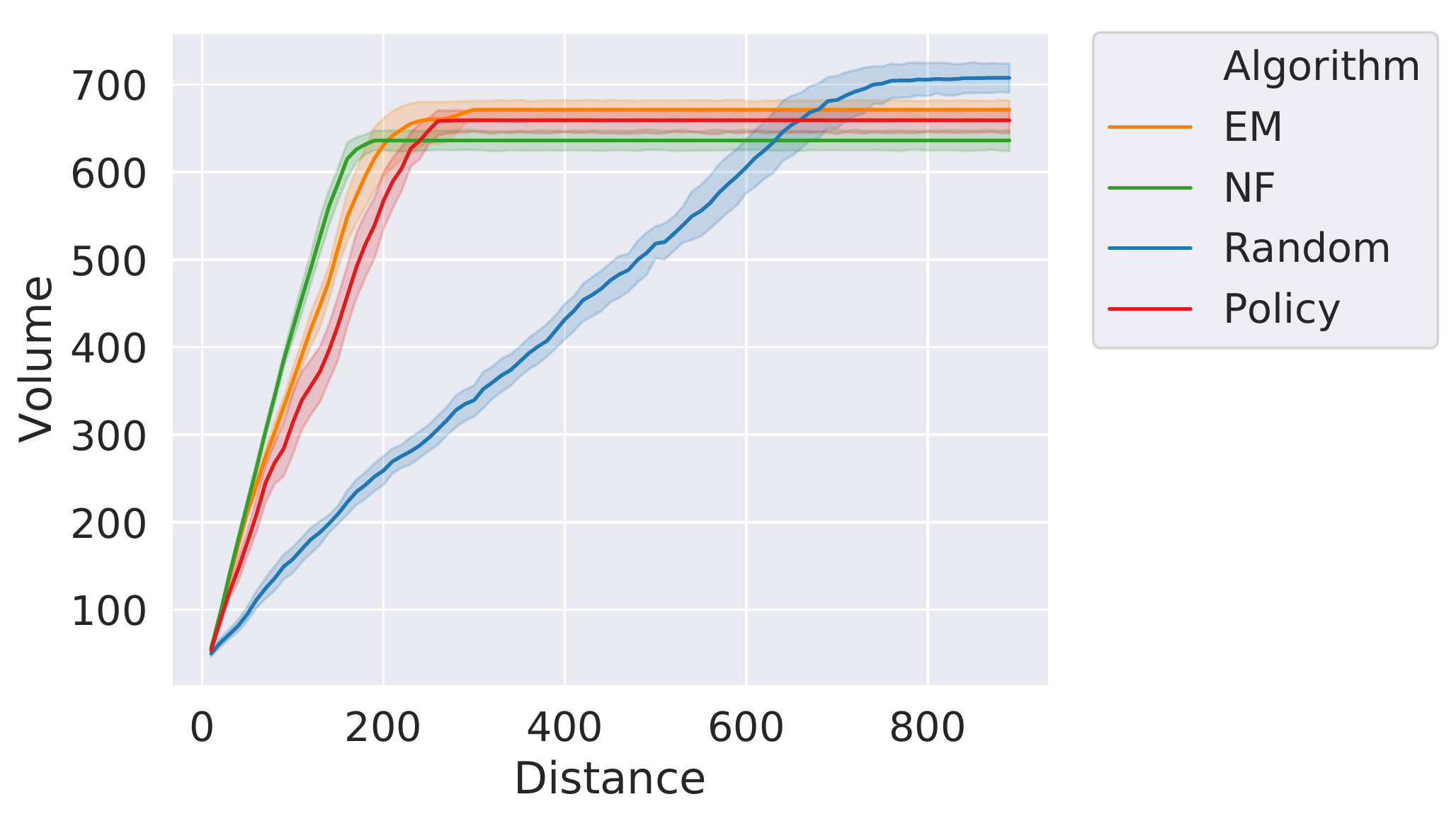}\label{test_right_volume}}
\vspace{-2mm}
\caption{Autonomous exploration results from our Gazebo simulation testing environments.}
\vspace{-5mm}
\end{figure*}

 We assume that an unmanned ground vehicle (UGV) must explore an indoor environment populated with 3D obstacles, it relies upon 6 degree-of-freedom SLAM, and it is permitted to reason about exploration with respect to the ground plane, where belief space planning is performed. The simulation environments used in this work are built in Gazebo and explored using the Robot Operating System (ROS). A simulated Clearpath Jackal robot is equipped with wheel odometry and a VLP-16 3D LiDAR. We restrict the sensor range of our LiDAR to 3 meters to induce challenging pose uncertainty in our exploration comparison. We adopt the default noise settings in Gazebo for the simulated sensors, and we manually add Gaussian noise to robot translation and rotation actions of standard deviation 0.01m and 0.08{\degree}, respectively. We use the same SLAM framework employed in \cite{Wang2019}, with the GTSAM library \cite{gtsam}. Dijkstra’s algorithm is used to generate a path to each frontier waypoint.

A 2D map of virtual landmarks %do not affect the SLAM process; they 
is maintained in the ground plane to provide the reward for each decision-making selection during the DRL policy training. The resolution of the map is 0.5m per cell and the standard deviation of the initial covariance of each virtual landmark is 0.2m. We terminate the exploration task once $85\%$ of the environment's ground plane has been explored. Additionally, we keep track of the robot's volumetric sensor coverage of the environment using a separate 3D occupancy map of the workspace.

Our graph neural network models are trained using PyTorch Geometric \cite{Fey2019}. The desktop used for policy training and simulation testing is equipped with an Intel i9 3.6Ghz CPU and an Nvidia Geforce Titan RTX GPU. For real-world exploration experiments (testing only), our algorithms run on a Dell Precision 3541 mobile workstation laptop which has an Intel Xeon CPU and an Nvidia Quadro P620 GPU.

%%%%%%%%%%%%%%%%%%%%%%%%%%%%%%%%%%%%%%%%%%%%%
\subsection{Policy Training}
\label{subsec:policy_training}

To evaluate the transferability of our proposed approach, we only use one environment during training. The office-like training environment is shown in Fig. \ref{train_env}. We randomly placed 8 objects in this environment. The robot has nine fixed initial locations in this environment from which it begins exploring.

We apply 15,000 training episodes in total, which, motivated by the expense of our ROS/Gazebo simulation, is orders of magnitude fewer than in our prior work with 2D landmark-based SLAM \cite{Chen2020}. We set the learning rate to 0.0001 and perform a policy update every 10 steps. The average reward obtained during training is shown in Fig. \ref{ave_reward}. Throughout this process, the state of the system is represented by the exploration graph, and the action space is comprised of the frontier nodes in the exploration graph. 

\begin{figure*}[t]
\centering
\subfigure[]{\includegraphics[width=0.47\columnwidth]{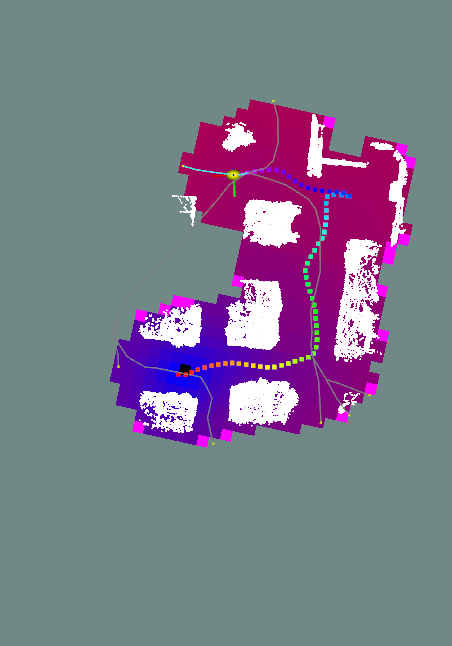}\label{test_real1}}
\subfigure[]{\includegraphics[width=0.47\columnwidth]{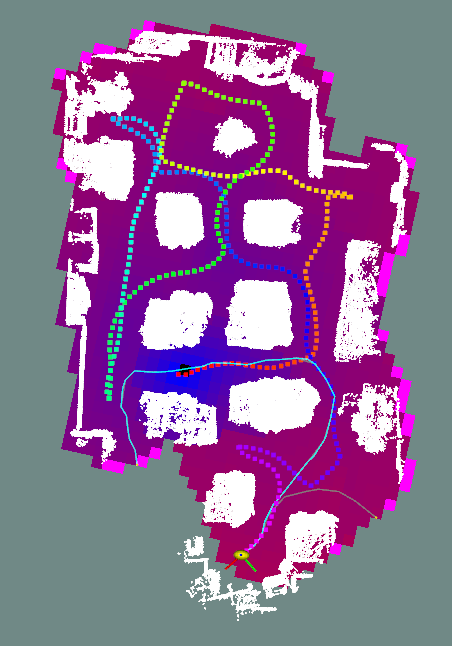}\label{test_real2}}
\subfigure[]{\includegraphics[width=0.47\columnwidth]{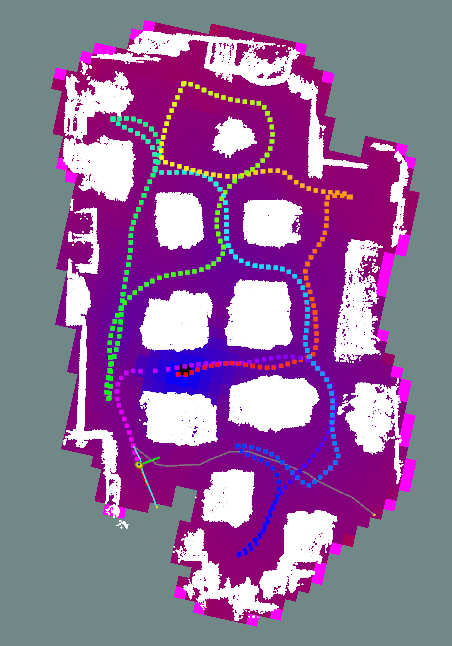}\label{test_real3}}
\subfigure[]{\includegraphics[width=0.47\columnwidth]{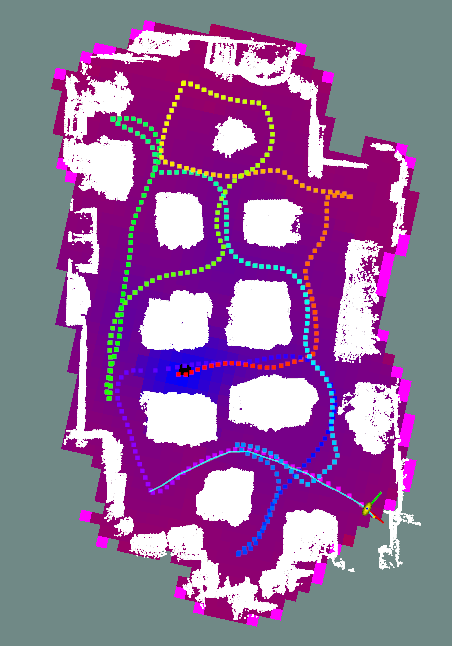}\label{test_real4}}
\caption{Ground-plane map and trajectory resulting from our graph-based policy's exploration of the Stevens ABS Engineering Center.}
\label{test_real}
\vspace{-6mm}
\end{figure*}

\begin{figure}[t]
\centering
\includegraphics[width=0.34\columnwidth]{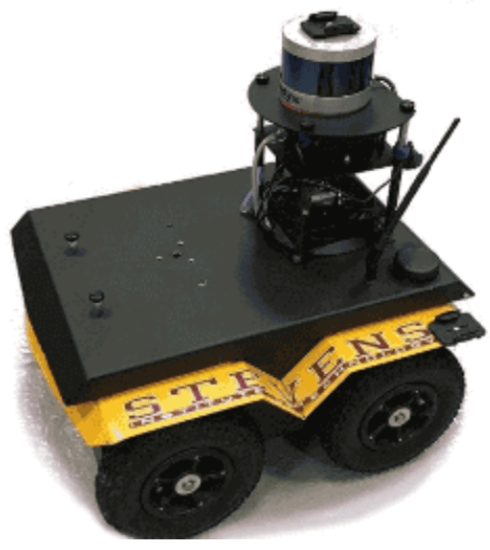}
\caption{The Clearpath Jackal UGV used in our experiments.}
\label{jackal}
\vspace{-0mm}
\end{figure}

\begin{figure}[t]
\centering
\includegraphics[width=0.64\columnwidth]{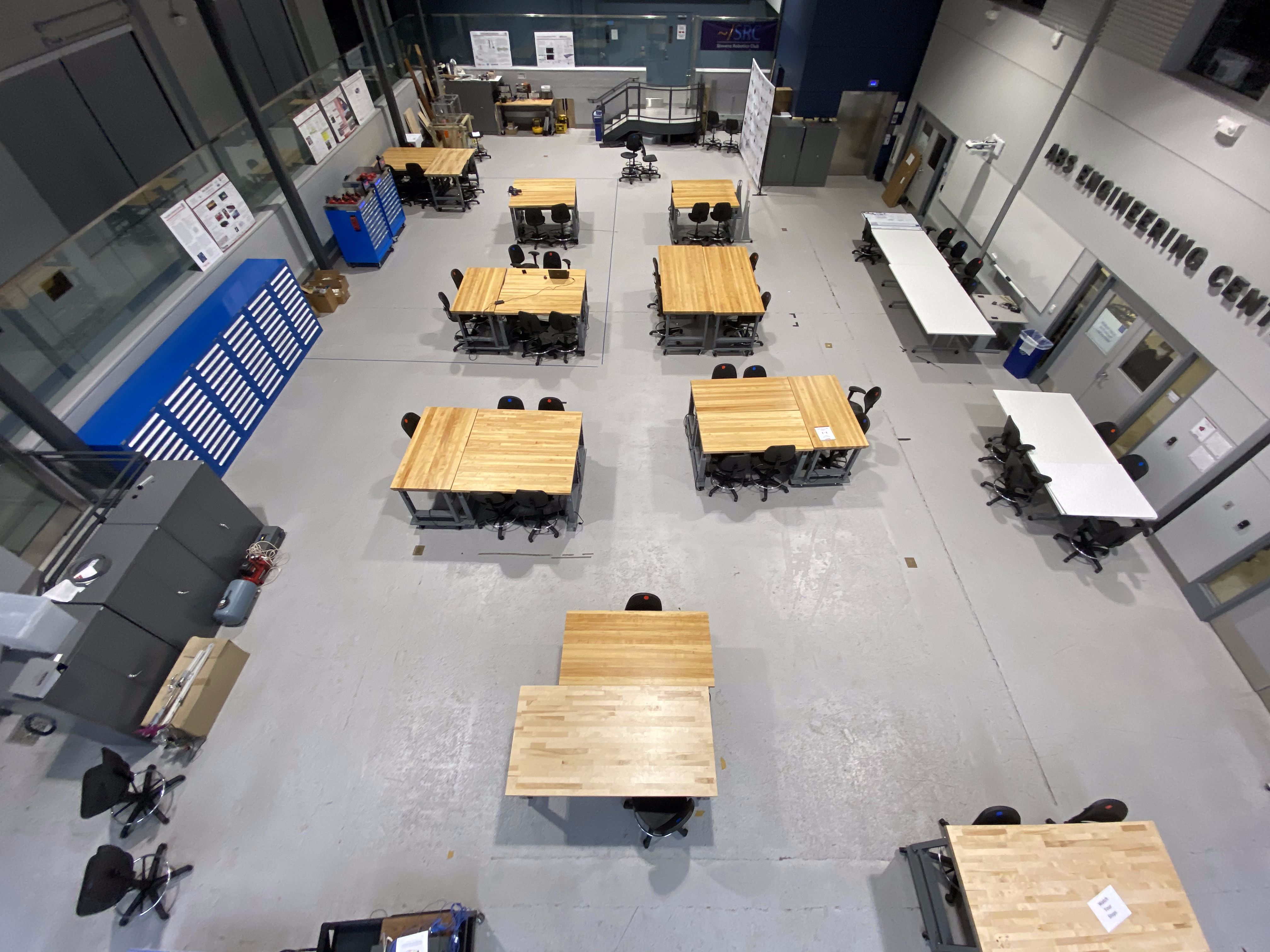}
\caption{Environment used for real-world deployment of our UGV. %(Stevens ABS engineering center)
}
\label{abs}
\vspace{-5mm}
\end{figure}
%%%%%%%%%%%%%%%%%%%%%%%%%%%%%%%%%%%%%%%%%%%%%
\subsection{Simulated Exploration Comparison}

During the test phase, the trained policy takes exploration graphs as input data to support the prediction of the next frontier waypoint. There is no uncertainty propagation at the robot's decision-making step, permitting real-time computation. We use the learned policy trained in the office-like environment to test in two different environments. In Fig. \ref{test_left_env}, we build a ``street environment" which has the same $20m \times 20m$ size as the training environment. However, this street environment contains fewer objects. Also, the size and the shape of these objects differ from the training environment. The exploration results are shown in Fig. \ref{test_left_error} and \ref{test_left_volume}. We compare the learned graph-based policy with (1) a nearest frontier (NF) approach \cite{Yamauchi1997}, (2) the EM algorithm \cite{Wang2017, Wang2019}, and (3) a random frontier selection approach over 10 exploration trials. For each trial, we always initialize the robot from the center of the testing environment. We compute the mean absolute error between the mapped 3D points associated with the estimated robot trajectories and the ground truth trajectories to determine the map error. We also compute the total 3D occupied volume mapped during each trial for the exploration efficiency comparison. 

The EM algorithm achieves the lowest map error in the end, but has a slightly worse exploration efficiency than the NF approach. Although the NF approach is the most efficient method for covering an unknown environment, it offers the worst map accuracy during exploration because it achieves the fewest loop closures. The random method achieves the most loop closures during exploration, but its long travel distances generate a large accumulated error that cannot be completely eliminated by these loop closures. It also covers the environment very inefficiently. %the second-worst final map error and the longest travel distance. 
Our learned policy achieves a very similar exploration efficiency to the EM algorithm, and its map accuracy is surpassed only by EM algorithm, whose performance we seek to emulate.

The second simulation testing environment is shown in Fig. \ref{test_right_env}. In this ``parking garage" environment, there are fifteen evenly spaced cars and the size of this environment is $30m \times 30m$, which is larger than the training environment. Like the first simulation environment, each exploration algorithm has 10 trials initialized from the center of the environment. The learned policy has a slightly worse exploration efficiency than the EM algorithm in this large environment, but it once again achieves the second lowest map error (again, second to the EM algorithm) compared with other exploration methods. All other algorithms have the same relative performance as in the ``street" environment.

%%%%%%%%%%%%%%%%%%%%%%%%%%%%%%%%%%%%%%%%%%%%%
\subsection{Real-world Exploration}

Our learned policy can also be transferred to a real-world environment. Our Clearpath Jackal UGV, shown in Fig. \ref{jackal}, is equipped with the same odometry and LiDAR sensing as the simulated UGV. We test our learned policy in the ABS Engineering Center at Stevens Institute of Technology shown in Fig. \ref{abs}. In Fig. \ref{test_real}, we present an example of autonomous exploration using the learned, graph-based policy. The robot pose is represented by red-green axes, and the yellow ellipsoid indicates the uncertainty of the current pose. The cyan path is the path to the selected frontier, and gray paths are for other unselected frontiers. The cost map covering the ground plane represents the uncertainty of the current virtual map, where blue color indicates the lowest uncertainty. Magenta represents the high uncertainty of our virtual map prior. We terminate the exploration task if there are no frontiers detected on the current map. In Fig. \ref{test_real1}, the robot chooses the nearest frontier to explore the environment, rather than the longer-distance path to obtain a loop closure, because this is the beginning of the exploration and a loop closure only reduces the uncertainty of a small portion of the current virtual map. In Fig. \ref{test_real2}, the robot selects a longer-distance path to obtain a loop closure to reduce the uncertainty of the current virtual map. In Fig. \ref{test_real3}, the uncertainty of the right bottom area on the map is reduced by taking the selected path. We present the final exploration result in Fig. \ref{test_real4}. The overall exploration process is shown in our video attachment.

%%%%%%%%%%%%%%%%%%%%%%%%%%%%%%%%%%%%%%%%%%%%%%%%%%%%%%%%%%%%%%%%%%%%%%%%%%%%%%%%%%%%%%%%%
%%%%%%%%%%%%%%%%%%%%%%%%%%%%%%%%%%%%%%%%%%%%%%%%%%%%%%%%%%%%%%%%%%%%%%%%%%%%%%%%%%%%%%%%%
%\vspace{-2mm}

\section{Conclusions}
\label{sec:conclusions}
In this paper, we present a zero-shot transfer learning framework for mobile robot exploration under uncertainty that leverages an exploration graph as an efficient abstraction of a robot's state and environment. We have enhanced the DRL GNN framework developed in our prior work \cite{Chen2020} so it can be applied, for the first time, to the exploration of environments populated with complex obstacles, perceived using dense 3D range observations. Successful training now depends on high-fidelity simulation, and accordingly, our approach offers a highly efficient training process to meet these requirements; the exploration policy is trained in a single virtual environment and is successfully transferred to both virtual and real environments containing different obstacle quantities, arrangements, and geometries.  %Also, we optimized the exploration graph in our previous work. 
The exploration graph %contains poses and candidate frontiers only, which has more generalizability and 
proposed in this work offers generality that is suitable for a wide variety of real-world exploration tasks, which we hope to study further in future work. %In the future, we will provide the information of the current environment (size, number of objects, etc.) to the exploration graph so 
Anticipated future versions of this system will adjust their exploration strategies to adapt to new environments. Also, 
although our policy is unlikely to exceed the EM algorithm's performance, because we adopt its utility function to assign rewards, we aim to enhance our reward function in future work to allow our framework to outperform the EM algorithm.

%%%%%%%%%%%%%%%%%%%%%%%%%%%%%%%%%%%%%%%%%%%%%%%%%%%%%%%%%%%%%%%%%%%%%%%%%%%%%%%%%%%%%%%%%
%%%%%%%%%%%%%%%%%%%%%%%%%%%%%%%%%%%%%%%%%%%%%%%%%%%%%%%%%%%%%%%%%%%%%%%%%%%%%%%%%%%%%%%%%
\vspace{2mm}

\section*{Acknowledgments}

This research has been supported by the National Science Foundation, grant numbers IIS-1652064 and IIS-1723996.

%\vspace{2mm}

%%%%%%%%%%%%%%%%%%%%%%%%%%%%%%%%%%%%%%%%%%%%%%%%%%%%%%%%%%%%%%%%%%%%%%%%%%%%%%%%%%%%%%%%%
%%%%%%%%%%%%%%%%%%%%%%%%%%%%%%%%%%%%%%%%%%%%%%%%%%%%%%%%%%%%%%%%%%%%%%%%%%%%%%%%%%%%%%%%%

\end{document}